\documentclass[letterpaper]{article} 
\usepackage{aaai2026}  
\usepackage{times}  
\usepackage{helvet}  
\usepackage{courier}  
\usepackage[hyphens]{url}  
\usepackage{graphicx} 
\usepackage{natbib}  
\usepackage{caption} 
\usepackage{algorithm}
\usepackage{algorithmic}

\usepackage{newfloat}
\usepackage{listings}
\DeclareCaptionStyle{ruled}{labelfont=normalfont,labelsep=colon,strut=off} 
\floatstyle{ruled}
\newfloat{listing}{tb}{lst}{}
\floatname{listing}{Listing}
\usepackage{amsmath}
\usepackage{amssymb}
\usepackage{booktabs}
\usepackage{makecell}
\usepackage{multirow}
\usepackage{romannum}
\usepackage{subcaption}
\usepackage{tabularray}

\nocopyright 

\title{FedSM: Robust Semantics-Guided Feature Mixup for Bias Reduction \\in Federated Learning with Long-Tail Data}
\author{
    Jingrui Zhang\textsuperscript{\rm 1, 2},
    Yimeng Xu\textsuperscript{\rm 1, 2},
    Shujie Li\textsuperscript{\rm 1, 3},
    Feng Liang\textsuperscript{\rm 1}\thanks{Corresponding authors.},
    Haihan Duan\textsuperscript{\rm 1},
    Yanjie Dong\textsuperscript{\rm 1},
    Victor C. M. Leung\textsuperscript{\rm 1},
    Xiping Hu\textsuperscript{\rm 1}\footnotemark[1]
}
\affiliations{
    \textsuperscript{\rm 1} Artificial Intelligence Research Institute, Shenzhen MSU-BIT University,\\
    \textsuperscript{\rm 2} School of Computer Science \& Technology, Beijing Institute of Technology,\\
    \textsuperscript{\rm 3} Department of Electrical and Electronic Engineering, The University of Hong Kong\\

    \footnotemark[1] \{fliang, huxp\}@smbu.edu.cn
}

\usepackage{bibentry}

\begin{document}

\maketitle

\begin{abstract}
Federated Learning (FL) enables collaborative model training across decentralized clients without sharing private data. 
However, FL suffers from biased global models due to non-IID and long-tail data distributions. 
We propose \textbf{FedSM}, a novel client-centric framework that mitigates this bias through semantics-guided feature mixup and lightweight classifier retraining. 
FedSM uses a pretrained image-text-aligned model to compute category-level semantic relevance, 
guiding the category selection of local features to mix-up with global prototypes to generate class-consistent pseudo-features.
These features correct classifier bias, especially when data are heavily skewed. 
To address the concern of potential domain shift between the pretrained model and the data, 
we propose probabilistic category selection, enhancing feature diversity 
to effectively mitigate biases. 
All computations are performed locally, requiring minimal server overhead. 
Extensive experiments on long-tail datasets with various imbalanced levels demonstrate that FedSM consistently outperforms state-of-the-art methods in accuracy, with high robustness to domain shift and computational efficiency.
\footnote{The source code is available at: https://github.com/DistriAI/FedSM.}
\end{abstract}


\section{Introduction}
\label{sec:intro}

The growing demand for big data utilization and privacy protection~\cite{shokri2015privacy} has driven the rise of federated learning (FL)~\cite{FL}, a distributed machine learning paradigm that enables multiple clients to collaboratively train a shared model without sharing raw data. 
Despite its promise, FL faces a fundamental challenge: data heterogeneity. 
Client datasets often differ significantly, leading to non-IID data distributions across clients. 
Moreover, real-world data typically follows a long-tail distribution, where a few head classes dominate in sample count, while many tail classes are severely underrepresented. 
This imbalance introduces bias into classifiers, skewing predictions toward head classes. 
As a result, poor client-side performance can degrade the global model, with client drift further amplifying the issue. 
Heterogeneity and long-tail distributions thus pose significant obstacles to effective FL.

\begin{figure}[!t]
    \centering
    \includegraphics[width=1.0\columnwidth]{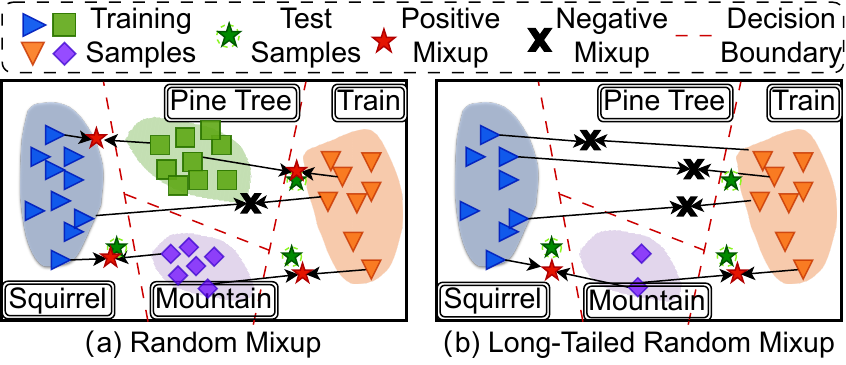}
    \caption{Illustration of Mixup problems. (a) Random Mixup ignores semantic relevance between categories and may blend unrelated samples, such as squirrel and train, across boundaries, producing synthetic data that misguides boundary refinement. (b) When the mountain category has significantly fewer samples or the pine tree category is absent, random Mixup has a higher chance of generating unrepresentative or even misleading synthetic samples.
}
    \label{fig:Intro_explain}
\end{figure}

In traditional centralized learning, resampling imbalanced datasets is a common strategy to improve classifier performance.  
Mixup~\cite{iclr2018mixup}, which interpolates between training samples to smooth decision boundaries, can enhance generalization by promoting feature continuity.  
However, when applied indiscriminately across unrelated categories, it may generate misleading pseudo-samples~\cite{selectiveMixup2024icml} or disrupt feature relationships~\cite{mixupRegularization2022JMLR}, potentially distorting decision boundaries rather than refining them.
This limitation is magnified in FL when data exhibit long-tail distribution, where global data diversity is fragmented across clients and clients may even lack samples from certain categories. 
Inappropriate mixes may exacerbate classification bias rather than alleviate it.
For example, as shown in Fig.~\ref{fig:Intro_explain}, images of squirrels often share background elements, like pine trees or mountains, with train images, even though squirrels and trains rarely co-occur. 
Mixing semantically related samples (e.g., squirrel and mountain in Fig.~\ref{fig:Intro_explain}(a)) can yield informative pseudo-samples that better align with test distributions, improving classification. 
In contrast, mixing semantically unrelated samples (e.g., squirrel and train) can distort decision boundaries.
This issues worsens under long-tail distributions or with missing categories, a setting that can be common in FL (Fig.~\ref{fig:Intro_explain}(b)).

To address the challenges of long-tail distributions in FL, we propose \textit{FedSM}, a \textit{S}emantic relevance-guided \textit{M}ixup method for feature augmentation that reduces classification bias. 
During mixup, head-class samples are used to enrich tail-class representations. 
A highlight of FedSM is that it generates more informative and robust pseudo-features for classifier retraining based on category pairs' semantic relevance provided by a pretrained image-text-aligned model (such as CLIP~\cite{clip}).
We first infer category relevance from label semantics using the model’s cross-modal reasoning capabilities. 
This semantic relevance is then used to guide sample pairing for mixup, ensuring that augmentation reflects meaningful, feature-level correlations rather than superficial similarity. 
Finally, each local classifier is retrained with the augmented data for a few rounds, improving its discriminative capability efficiently.
A major concern is the potential error caused by the domain shift between the pretrained model and the target data.
To improve FedSM's robustness to domain shift, we introduce a probability-based selection strategy that promotes diversity in sample selection. 
Furthermore, to explore a practical approach to refine the pretrained model for the domain shift issue, 
we try to fine-tune the pretrained image encoder to help quickly adapt to out-of-domain training images.
Both efforts are demonstrated to be effective in different datasets and data imbalance levels.
Our main \textbf{contributions} are as follows:
\begin{itemize}
    \item We propose FedSM, a novel semantics-guided mixup strategy for FL with long-tail data. 
    It selects semantic relevant category pairs to generate balanced pseudo-features to mitigate classifier bias.
    \item FedSm is robust to domain shifts.
    The probabilistic category pair selection approach enhances feature diversity, while fine-tuning the pretrained image encoder allows for further feature space refinement.
    \item FedSM is a lightweight and efficient client-centric framework. 
    All computations performed locally, minimizing server overhead.
    \item Extensive experiments demonstrate that FedSM consistently outperforms state-of-the-art methods in accuracy, robustness to domain shift, and computational efficiency. 
\end{itemize}

\section{Related Work}
\label{sec:related_work}

\textbf{Long-Tail Learning:}
Two primary strategies dominate long-tail learning: \textit{re-weighting} and \textit{decoupled retraining}.  
Re-weighting assigns varying weights to samples based on category frequency, increasing the emphasis on tail classes to counterbalance head-class dominance.  
For example, Cui~\textit{et~al.}~\citeyear{effnumber} proposed an exponential weighting method to redistribute importance across categories.  
Similarly, AREA~\cite{area} recalibrates classifier updates by estimating the effective area in the feature space.
Decoupled retraining, in contrast, separates feature learning and classifier learning.  
Kang~\textit{et~al.}~\citeyear{kang2019decoupling} introduced a two-stage training pipeline, feature learning and classifier learning, to learn balanced classifiers.  
BBN~\cite{zhou2020bbn} further evolved this into a dual-branch architecture with shared parameters, one branch for standard training and the other for classifier refinement.  
While effective, these approaches are designed for centralized settings and do not directly translate to FL, where decentralized data introduces additional challenges.

\textbf{Federated Learning with Heterogeneous Data:}
Most existing works address client-level heterogeneity in FL but often assume class distributions are uniform, overlooking global class imbalance.
Solutions typically fall into two categories:  
server-side methods that mitigate the impact of heterogeneity~\cite{chen2021fedbe}, and  
methods that preserve consistency between local and global models~\cite{huang2021behavior, MLSYS2020_38af8613,fedpall_iccv25}.  
For example, CCVR~\cite{ccvr} retrains classifiers using virtual features sampled from a Gaussian Mixture Model to address heterogeneity, though its performance deteriorates under long-tail distributions.  
Other methods focus on client selection for data complementarity~\cite{yang2021federated, duan2020self}, often requiring revealing local data distribution, undermining FL's privacy guarantees. 
Our method is applicable to global class heterogeneity and requires retraining the aggregated classifier with locally augmented data. 

\textbf{Federated Learning with Long-Tail Data:}
When long-tail data is distributed across clients, local models often develop severely biased representations. 
CReFF~\cite{CReFF} and CLIP2FL~\cite{clip2fl} upload averaged gradients of local data to the server, which then synthesizes balanced pseudo-features for classifier retraining. Although the averaging process is non-invertible, it still raises potential privacy concerns.  
RUCR~\cite{rucr} employs a Mixup-inspired strategy~\cite{iclr2018mixup} to generate pseudo-features.
Unlike these methods, FedSM adopts a client-centric approach and avoids server-side gradient sharing and privacy risks. 
It leverages semantic guidance from image-text models to generate pseudo-features locally, enabling robust and privacy-preserving augmentation for long-tail FL scenarios.

\section{Method}
\subsection{Problem Setting}

We assume a standard FL setup with $K$ clients holding non-IID, long-tail data. 
The goal is to train a shared feature extractor and classifier that generalizes well despite client drift and label imbalance.
Let $\mathcal{D}^k$ denote the local dataset on client $k$, with size $n^k = |\mathcal{D}^k|$.  
The global dataset is defined as $\mathcal{D} = \bigcup_{k=1}^K \mathcal{D}^k$ and consists of $C$ classes.
For class $c$, let $\mathcal{D}_c^k = \{(x, y) \in \mathcal{D}^k \mid y = c\}$ be the subset of samples distributed to client $k$, and $n_c^k = |\mathcal{D}_c^k|$ its size.  
The total number of samples in class $c$ across all clients is $N_c = \sum_{k=1}^K n_c^k$.  
The global data follows a long-tail distribution, i.e., when sorted by class frequency such that $N_1 \geq N_2 \geq \dots \geq N_C$, we have $N_1 \gg N_C$.

The standard FL process involves:  
1) The server broadcasts the global model to clients;  
2) Clients update local models using private data;  
3) Locally updated models are sent back to the server for aggregation;  
This cycle repeats until convergence.
Our objective is to learn a high-performance global model for image classification under the constraint of long-tail distributed data in the FL setting.

\begin{figure}[!t]
    \centering
    \includegraphics[width=1 \columnwidth]{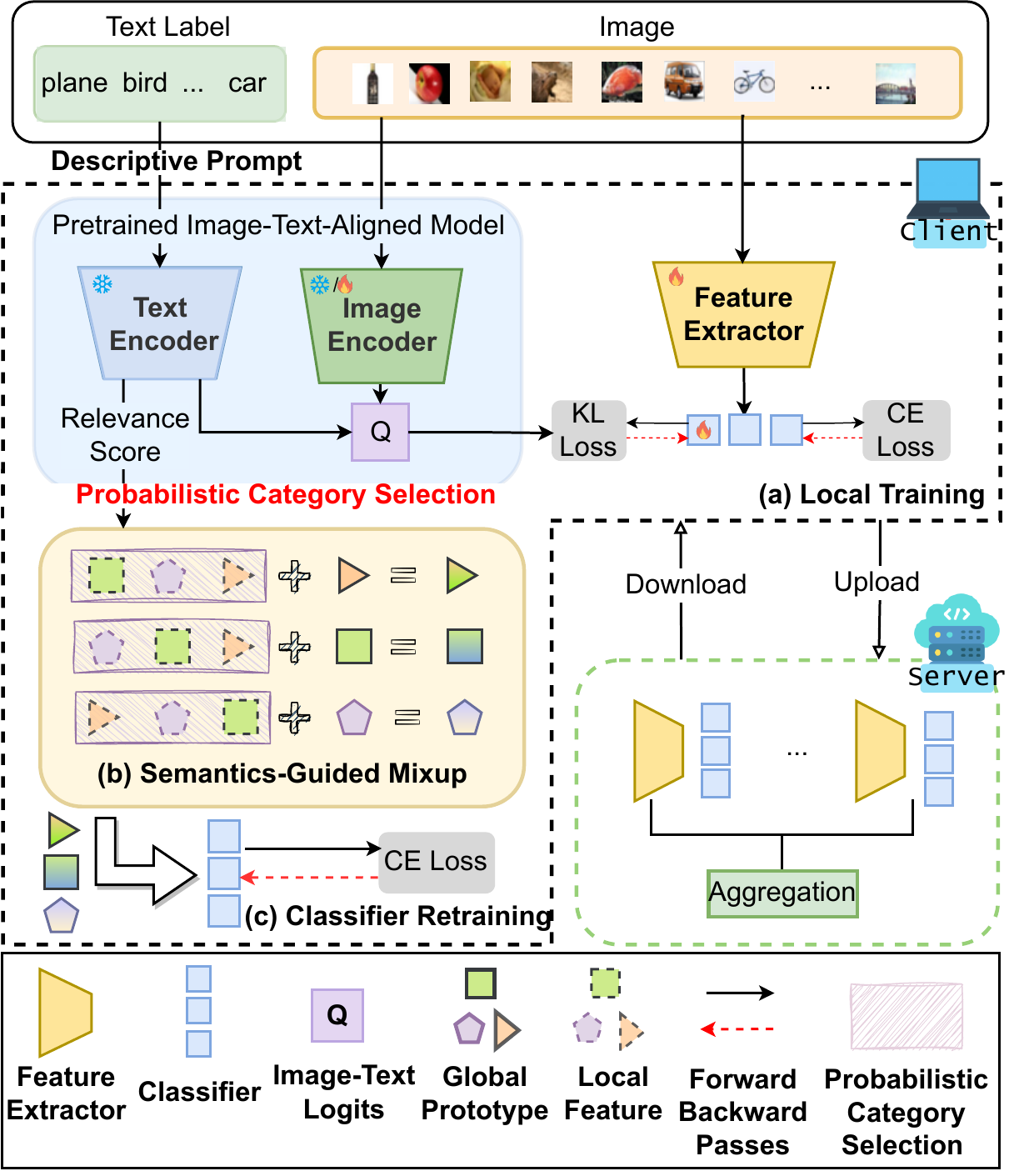}
    \caption{Overview of the FedSM framework. The client side consists of three key phases: a) local training, b) label relevance-guided feature mixup, and c) classifier retraining. 
    }
    \label{fig:framework}
\end{figure}

\subsection{Overview of Our Framework}
FedSM follows the standard federated learning process:  
1) The server distributes the global model to each client;
2) Clients update their local models using private data;  
3) Clients send the updated models back, and the server aggregates them.  
These steps repeat until convergence.
This study primarily focus on the client-side training with three phases: local training, label relevance-guided feature mixup, and classifier retraining. 
As shown in Fig.~\ref{fig:framework}, the local training begins with knowledge distillation from a pretrained image-text-aligned model, enhancing the representational capacity of the local feature extractor.
Next, balanced pseudo-features are generated locally for classifier retraining. 
We use text features from the image-text-aligned model to compute semantic relevance between labels for mixup pairing, based on which we select samples for mixup.
This relevance-guided mixup ensures that generated pseudo-features remain semantically consistent and do not overlap with unrelated classes.  
Motivated by prior work~\cite{CReFF} showing that classification bias mainly stems from the classifier rather than the feature extractor $f$, we retrain only the classifier $g$ after several local training rounds to correct these biases.
The server side remains unchanged, executing standard FL procedures without any additional modifications.

\subsection{Local Training}
In the local training phase, our goal is to enhance the model's representation capability of aligning image features with text semantics by transferring knowledge from a pretrained image-text-aligned model.  
To this end, we adopt a knowledge distillation strategy within a teacher–student framework, where the image-text-aligned model serves as the teacher, guiding the local model (student) during learning.
This image-text-aligned model is required to have strong semantic understanding of both visual and textual modalities with two encoders: an image encoder $Enc_I$ and a text encoder $Enc_T$.  
Given an input image $x$ and its corresponding text label $l$, we compute the visual feature as 
$h_v = Linear(Enc_I(x)) \in \mathbb{R}^d$ and the text feature as $h_t = Linear'(Enc_T(l)) \in \mathbb{R}^d$, where $d$ is the feature dimension.
The image-text-aligned model output logits are calculated as:
\[
q = [\langle h_v, h_{t_1} \rangle, \langle h_v, h_{t_2} \rangle, \ldots, \langle h_v, h_{t_C} \rangle],
\]
where $\langle \cdot, \cdot \rangle$ denotes cosine similarity between visual and textual features across all $C$ categories.
In client $k$, The local model prediction $p = w^k(x)$ is obtained by forwarding $x$ through the local model $w^k$. The total training loss combines supervised and distillation objectives:
\begin{equation}
    \label{loss local learning}
    L = L_{\text{CE}}(y, p) + L_{\text{KL}}(q \parallel p){}
\end{equation}
where $y$ is the category label of $p$, $L_{\text{CE}}$ is the cross-entropy loss, and $L_{\text{KL}}$ denotes the Kullback–Leibler divergence~\cite{joyce2011KL}.

After local updates in the $t$-th round, clients in a randomly selected subset $U^t$ upload their models ${w}^k$ to the server.  
Following standard FL aggregation, the server computes the updated global model as a weighted average of client models, 
given by:
\begin{equation}
    \label{eq:update_global_model}
    w = \sum_{k \in U^t} \frac{|\mathcal{D}^k|}{\sum_{k \in U^t} |\mathcal{D}^k|} \, w^k.
\end{equation}

\subsection{Image Feature Mixup Guided by Label Relevance}
\textbf{Feature mixup.}  
Sample-level augmentation techniques such as MixUp~\cite{iclr2018mixup} and CutMix~\cite{yun2019cutmix} are simple yet effective for mitigating long-tail distributions.  
However, these methods originally operate at the pixel level and do not exploit higher-level feature-space mixing, limiting their applicability in FL, where decentralized data and privacy constraints make raw image sharing impractical.
To overcome this limitation, FedSM performs mixup in the feature space, leveraging both global category prototypes and local features.  
This approach maintains a global perspective to reduce bias while preserving client-specific characteristics and adhering to FL privacy principles.

The global prototype for category $c$ is the aggregation of local category prototypes from all clients, which is defined as:
\begin{equation}
    \begin{gathered}
        z^{\text{global}}_c = \frac{1}{N_c} \sum_{k=1}^{K} f_c^k \cdot n_c^k, \quad
        f_c^k = \frac{1}{n_c^k} \sum_{i=1}^{n_c^k} g(x_{c,i}^k),
    \end{gathered}
    \label{eq:get_global_prototype}
\end{equation}
where $g(\cdot)$ is the local feature extractor, $x_{c,i}^k$ is the $i$-th sample of category $c$ on client $k$, and $f_c^k$ is the client-level category prototype uploaded to the server.

A pseudo feature $r^k_c$ for category $c$ on client $k$ is generated by mixing the global prototype of category $c$ with a local feature from the most semantically relevant category $v$:
\begin{equation}
    r^k_c = (1 - \lambda) \cdot z^k_v + \lambda \cdot z^{\text{global}}_c,
    \label{eq:mix}
\end{equation}
where $z^k_v$ is a local feature of category $v$, and $\lambda$ is a mixup coefficient that balances the importance of generalization (global prototype) and personalization (local feature).

\textbf{Category relevance estimation.}  
This step selects the most semantically relevant category $v$ for a target category $c$.  
Unlike prior methods that rely on co-occurrence or feature similarity, FedSM leverages label semantics via a pretrained image-text-aligned model.

Specifically, FedSM uses the model’s text encoder to estimate the similarity between categories based solely on their textual labels.  
Each label $l_i$ is converted into a descriptive $\text{phrase}_i$ (e.g., ``a photo of \{label\}'').
The semantic relevance score $\alpha_{i,j}$ between categories $i$ and $j$ is computed as:
\begin{equation}
    \alpha_{i,j} = Nonlinear (\langle Enc_T(\text{phrase}_i), Enc_T(\text{phrase}_j) \rangle),
    \label{eq:relevance_score}
\end{equation}
where $\langle \cdot, \cdot \rangle$ denotes similarity between encoded text features, default to cosine, and $Nonlinear$ refers to an optional transformation (e.g., softmax).

The resulting relevance score $\alpha_{i,j}$ is interpreted as the probability of selecting a local feature $z^k_v$ from category $v$ in Eq.~\ref{eq:mix}, ensuring semantic consistency in pair selection.  
In FedSM, each client ranks its available categories based on relevance scores and assigns higher selection probabilities to more relevant categories, promoting semantically meaningful mixup and generating balanced pseudo data.

This probabilistic strategy offers two key advantages:  
1) It mitigates domain shift between the pretrained model and the downstream FL task by introducing controlled randomness, reducing over-reliance on potentially misaligned semantic priors.  
2) It enhances mixup diversity and robustness by allowing feature synthesis from a broader pool of relevant categories, especially beneficial when the top-matching categories are absent from a client's local dataset.

Moreover, applying a nonlinear transformation to the similarity scores allows fine-grained control over the distribution sharpness, amplifying confidence in top choices or smoothing across multiple candidates, further improving flexibility and stability in pseudo feature generation.

Each client generates $S$ semantics-guided pseudo-features per category.  
Let $r^k_{c,i}$ denote the $i$-th pseudo feature for category $c$ on client $k$, then the complete pseudo feature set on client $k$ is defined as:
\begin{equation}
    \mathcal{R}^{k} = \{ r^k_{c,i} \mid c \in C,\, i = 1, \ldots, S \}.
\end{equation}

\textbf{Classifier Retraining.}  
After local training, client $k$ refines its classifier $g^k$ using the generated pseudo feature set $\mathcal{R}^k$.  
This retraining step aims to mitigate classification bias and further enhance robustness to domain shift by leveraging semantically enriched, balanced synthetic data.  
The loss function is the cross entropy loss:
\begin{equation}
    L_{\text{CE}}(g^k; \mathcal{R}^k) = \frac{1}{|\mathcal{R}^k|} \sum_{(r, y) \in \mathcal{R}^k} -y \log\big(\sigma(g^k(r))\big),
\end{equation}
where $\sigma$ denotes the softmax function. 

This lightweight retraining phase is performed only on the classifier $g^k$, making it computationally efficient while improving model generalization.  
The final model $w^k = \{f^k, g^k\}$, consisting of the locally updated feature extractor and the calibrated classifier, is then uploaded to the server for aggregation.
Unlike prior methods~\cite{CReFF, clip2fl} that retrain local models at every FL communication round, FedSM performs classifier retraining only in the final few rounds, significantly reducing computational overhead.

Overall, the global model is iteratively updated through client-side local training and classifier retraining with semantics-guided mixed-up features, as outlined in Algorithm~\ref{alg:pseudo-code}.

\begin{algorithm}[!t]
    \caption{FedSM Training at Communication Round $t$}
    \label{alg:pseudo-code}
    \renewcommand{\algorithmicrequire}{\textbf{Input:}}
    \renewcommand{\algorithmicensure}{\textbf{Output:}}

    \begin{algorithmic}[1]
        \REQUIRE Global model $w^t = \{f^t, g^t\}$
        \ENSURE Updated global model $w^{t+1} = \{f^{t+1}, g^{t+1}\}$

        \STATE \textbf{Server-side:}
        \STATE Randomly sample a set of online clients $U^t$
        \STATE Send global model $w^t$ to all $k \in U^t$

        \STATE \textbf{Client-side (for each $k \in U^t$):}
        \STATE Update local model using Eq.~\ref{loss local learning}\\
        /* \textit{generate pseudo-features for classifier retraining} */
        \IF{$t \geq \texttt{total\_rounds} - \texttt{retraining\_rounds}$}
            \STATE Compute category relevance via Eq.~\ref{eq:relevance_score}
            \STATE Obtain global prototypes via Eq.~\ref{eq:get_global_prototype}
            \STATE Generate pseudo feature set via Eq.~\ref{eq:mix}
            \STATE Retrain classifier using pseudo-features
            \STATE Set local model $w_k^{t+1} = \{f_k^{t+1}, g_k^{t+1}\}$
        \ENDIF

        \STATE Send updated local model $w_k^{t+1}$ to server

        \STATE \textbf{Server-side:}
        \STATE Aggregate received models via Eq.~\ref{eq:update_global_model}
    \end{algorithmic}
\end{algorithm}

\setlength{\tabcolsep}{1mm}

\begin{table*}[!t]
\fontsize{9}{10}\selectfont
    \centering
    \begin{tabular}{cccccccc}
    \toprule
        \multirow{2}{*}{\textbf{Type}} & \multirow{2}{*}{\textbf{Method}} & \multicolumn{3}{c}{\textbf{CIFAR-10-LT}} & \multicolumn{3}{c}{\textbf{CIFAR-100-LT}}  \\ \cline{3-8} 
                &   & IF=100    & IF=50     & IF=10     & IF=100    & IF=50     & IF=10 \\ \midrule
            \multirow{7}{*}{ \makecell[c]{\Romannum{1}}} 
                & FedAvg~\cite{fedavg} &  57.3 $\pm$ 1.7  &  61.0 $\pm$ 3.6  &  72.0 $\pm$ 3.6  &  31.6 $\pm$ 0.7  &  35.9 $\pm$ 0.3  &  47.6 $\pm$ 0.8  \\
                & FedAvgM~\cite{fedavgm} &  56.7 $\pm$ 1.6  &  61.2 $\pm$ 4.0  &  71.9 $\pm$ 4.0  &  31.7 $\pm$ 0.7  &  36.3 $\pm$ 0.5  &  47.3 $\pm$ 0.9  \\
                & FedProx~\cite{fedprox} &  54.4 $\pm$ 2.2  &  60.4 $\pm$ 2.5  &  69.8 $\pm$ 2.9  &  30.4 $\pm$ 0.4  &  34.3 $\pm$ 0.4  &  43.9 $\pm$ 0.4   \\
                & FedNova~\cite{fednova} &  56.5 $\pm$ 1.6  &  61.0 $\pm$ 4.4  &  72.6 $\pm$ 5.1  &  31.6 $\pm$ 1.0  &  36.1 $\pm$ 0.3  &  47.5 $\pm$ 0.6   \\
                & CCVR~\cite{ccvr} & 60.4 $\pm$ 2.2  & 68.2 $\pm$ 2.0  & 74.4 $\pm$ 2.3  & 25.1 $\pm$ 0.9  & 27.1 $\pm$ 2.0  & 36.0 $\pm$ 1.0  \\
                & MOON~\cite{moon} &  57.5 $\pm$ 1.1  &  61.6 $\pm$ 3.6  &  73.0 $\pm$ 3.2  &  31.9 $\pm$ 0.9  &  36.1 $\pm$ 0.3  &  47.5 $\pm$ 0.8  \\
            \midrule
            \multirow{2}{*}{\makecell[c]{\Romannum{2}} }
                & Fed-Focal~\cite{fedfocalloss} & 52.9 $\pm$ 1.9  & 58.1 $\pm$ 2.6  & 74.9 $\pm$ 5.5  & 30.3 $\pm$ 0.7  & 34.6 $\pm$ 0.9  & 41.4 $\pm$ 0.8  \\
                & RatioLoss~\cite{fedratioloss} & 56.0 $\pm$ 2.2  & 65.0 $\pm$ 2.7  & 72.8 $\pm$ 5.4  & 31.7 $\pm$ 0.9  & 34.7 $\pm$ 0.9  & 42.6 $\pm$ 1.1   \\
            \midrule
            \multirow{3}{*}{\makecell[c]{\Romannum{3}}}  
                & CReFF~\cite{CReFF} & 69.9 $\pm$ 1.2  & 72.6 $\pm$ 1.1  & 79.6 $\pm$ 1.5  & 26.9 $\pm$ 0.7  & 30.3 $\pm$ 0.6  & 37.8 $\pm$ 1.0  \\
                & RUCR~\cite{rucr}   & 61.3 $\pm$ 0.8  & 65.1 $\pm$ 3.4   & 79.3 $\pm$ 1.2  & 33.7 $\pm$ 0.1  & 37.4 $\pm$ 0.0  & 48.8 $\pm$ 0.2  \\ 
                & CLIP2FL~\cite{clip2fl} & \underline{71.2 $\pm$ 0.8}  & \underline{72.6 $\pm$ 1.8}  & 80.7 $\pm$ 1.7  & \underline{36.0 $\pm$ 0.7}  & \underline{39.6 $\pm$ 0.6}  & 47.2 $\pm$ 0.5  \\
                \midrule
              & \textbf{FedSM+MetaCLIP (Ours)}  & 70.4 $\pm$ 0.7  & 71.6 $\pm$ 0.9 & \underline{80.9 $\pm$ 1.1}  & 35.6 $\pm$ 0.7 & 39.5 $\pm$ 0.5  & \underline{50.2 $\pm$ 0.8}  \\ 
              & \textbf{FedSM+CLIP (Ours)}  & \textbf{72.2 $\pm$ 0.9}  & \textbf{74.4 $\pm$ 1.0}  & \textbf{82.2 $\pm$ 0.5}  & \textbf{37.8 $\pm$ 0.5}  & \textbf{41.2 $\pm$ 0.4}  & \textbf{50.7 $\pm$ 0.7}  \\ 
        \bottomrule
        
    \end{tabular}
        \caption{Top-1 accuracy(\%) of different FL algorithms on the CIFAR-10-LT and CIFAR-100-LT datasets. ``\Romannum{1}'', ``\Romannum{2}'', and ``\Romannum{3}'' represent types of heterogeneity-oriented, imbalance-oriented, and heterogeneity and imbalance-oriented, respectively.}
    \label{table:top_one_acc_cifar}
\end{table*}

\section{Evaluation}
\subsection{Experimental Setup and Implementation Details}
\textbf{Datasets.}  
We evaluate FedSM on three long-tail benchmarks: CIFAR-10-LT~\cite{CReFF}, CIFAR-100-LT~\cite{CReFF}, and ImageNet-LT~\cite{iamgenet-lt}.  
CIFAR-10-LT and CIFAR-100-LT are derived from CIFAR-10 and CIFAR-100~\cite{cifar}, respectively, by sampling with varying imbalance factors (IF): 100, 50, and 10.  
An imbalance factor of $100$ means the most frequent class has 100 times more samples than the least frequent one.  
To simulate non-IID data across clients, we adopt Dirichlet partitioning with $\alpha = 0.5$, following CReFF~\cite{CReFF}.
ImageNet-LT is a long-tail subset of ImageNet~\cite{ImageNet}, containing 115.8K images across 1000 categories.  
It has a predefined distribution with up to 1280 images in head classes and as few as five in tail classes.  
For ImageNet-LT, we use Dirichlet partitioning with $\alpha = 0.1$ to introduce data heterogeneity among clients.

\textbf{Implementation and Setup.}  
For CIFAR-10-LT and CIFAR-100-LT, we use ResNet-8~\cite{resnet} as the feature extractor, and for the larger ImageNet-LT dataset, we adopt ResNet-50~\cite{resnet}.  
We use CLIP~\cite{clip} or MetaCLIP~\cite{metaclip_iclr24} as the image-text-aligned model. 
These models have been pretrained on rich image and text data from diverse domains and can be used to verify FedSM's performance under domain shifts. 
To align with the image-text-aligned model, a projection layer is added atop the base model to match the feature dimension. Both its text and image encoders are frozen during training.  
For the CLIP image encoder, we use the ViT-B/32 variant, consistent with the setup in CLIP2FL~\cite{clip2fl}.
CLIP is the default choice for other experiments if the image-text-aligned model is not specifically mentioned.
FedSM and other baseline methods are implemented within the FLGO framework~\cite{wangzheng2021federated,wangzheng2023flgo} relying on PyTorch.  
Each experiment is repeated five times with different random seeds for CIFAR-10-LT and CIFAR-100-LT, and three times for ImageNet-LT.  
All experiments are run on a single node equipped with four NVIDIA A800 GPUs.

\textbf{Training.}  
By default, we simulate 20 clients, with 40\% randomly selected for participation in each communication round.  
The classifier is retrained using 100 pseudo-features per class, following the common practice in recent works~\cite{clip2fl, CReFF}.  
We use the standard cross-entropy loss and run totally 200 communication rounds with 10 epochs per round.
The baseline methods~\cite{CReFF, clip2fl} retrain local models at every round, while FedSM performs classifier retraining with pseudo-features only in the final 50 rounds.
All experiments use Stochastic Gradient Descent (SGD) with a learning rate of 0.1 for local training and 0.01 for classifier retraining.  
The mixup coefficient $\lambda$ in Eq.~\ref{eq:get_global_prototype} is chosen randomly from range 0.65 to 0.90 and batch size is 32 across all datasets.

\begin{table*}[!t]
\fontsize{9}{10}\selectfont
    \centering
    \begin{tabular}{cccccc}
    \toprule
        \multirow{2}{*}{\textbf{Type}} & \multirow{2}{*}{\textbf{Method}} & \multirow{2}{*}{\textbf{Overall}} & \multicolumn{3}{c}{Divided Categories} \cr \cline{4-6} 
                &   &     & Many     & Medium & Few \\ \midrule
            \multirow{5}{*}{ \makecell[c]{Heterogeneity- \\ oriented}} 
                & FedAvg~\cite{fedavg} & 23.0 $\pm$ 2.0 & 34.9 $\pm$ 1.2 & 19.1 $\pm$ 1.0 & 7.0 $\pm$ 1.3 \\
                & FedAvgM~\cite{fedavgm} & 22.5 $\pm$ 2.2 & 33.9 $\pm$ 1.4 & 18.7 $\pm$ 1.4 & 6.0 $\pm$ 1.2 \\
                & FedProx~\cite{fedprox} & 22.9 $\pm$ 1.6 & 35.0 $\pm$ 1.8 & 17.1 $\pm$ 1.2 & 7.0 $\pm$ 0.9 \\
                & FedNova~\cite{fednova} & 24.7 $\pm$ 2.0 & 35.4 $\pm$ 0.8 & 20.6 $\pm$ 1.6 & 11.6 $\pm$ 0.5 \\
                & CCVR~\cite{ccvr}  & 25.7 $\pm$ 1.5 & 36.8 $\pm$ 1.5 & 20.6 $\pm$ 1.6 & 10.0 $\pm$ 0.9 \\
                & MOON~\cite{moon} & 24.1 $\pm$ 1.1 & 34.7 $\pm$ 0.5 & 20.4 $\pm$ 0.9 & 9.9 $\pm$ 1.2 \\
            \midrule
            \multirow{2}{*}{\makecell[c]{Imbalance- \\ oriented} }
                & Fed-Focal~\cite{fedfocalloss} & 21.5$\pm$ 1.8 & 31.0 $\pm$ 1.6  &  15.8 $\pm$ 1.6 & 6.8 $\pm$ 1.3  \\
                & RatioLoss~\cite{fedratioloss}  &  25.0 $\pm$ 3.0 & 35.9 $\pm$ 2.3  & 18.9 $\pm$ 1.9  & 7.4 $\pm$ 1.4  \\
            \midrule
            \multirow{2}{*}{\makecell[c]{Heterogeneity and \\ Imbalanced}}  
                & CReFF~\cite{CReFF} & 19.7 $\pm$ 1.5  & 34.8 $\pm$ 2.1  &  18.7 $\pm$ 1.8  &  8.3 $\pm$ 0.7  \\
                & CLIP2FL~\cite{clip2fl} & 27.5 $\pm$ 1.0  & 35.7 $\pm$ 2.1  & 26.9 $\pm$ 1.8  & \textbf{23.4 $\pm$ 1.4} \\
                \midrule
              & \textbf{FedSM+MetaCLIP (Ours)}  & \underline{29.3 $\pm$ 0.4} & \underline{37.0 $\pm$ 0.6} & \textbf{28.4 $\pm$ 1.5} & 22.1 $\pm$ 1.3 \\ 
              & \textbf{FedSM+CLIP (Ours)}  & \textbf{30.9 $\pm$ 0.2} & \textbf{38.0 $\pm$ 0.3} & \underline{27.4 $\pm$ 0.1} & \underline{23.0 $\pm$ 0.2} \\ 
        \bottomrule
        
    \end{tabular}
    \caption{Top-1 accuracy(\%) of different federated learning algorithms on the ImageNet-LT. }
    \label{table:top_one_acc_imagenet}
\end{table*}

\subsection{Results}
We compare FedSM against a range of FL algorithms that address data heterogeneity at varying levels.  
General approaches~\cite{fedavg, fedavgm, fedprox, fednova, ccvr, moon} target standard heterogeneous settings,  
while others~\cite{fedfocalloss, fedratioloss} specifically focus on class imbalance.  
The most relevant to our work are recent state-of-the-art (SOTA) methods~\cite{CReFF, clip2fl, rucr} designed for FL with long-tail data.

\paragraph{Results on CIFAR-10/100-LT.}  
Table~\ref{table:top_one_acc_cifar} reports the classification accuracy of various FL algorithms on CIFAR-10-LT and CIFAR-100-LT.
FedSM with CLIP consistently outperforms all baselines across different IFs, with performance improvement ranging from 1.0 to 1.9 percentage points compared to second best results.
Performance gains on CIFAR-100-LT are generally slightly higher than on CIFAR-10-LT.
A possible reason is that CIFAR-100-LT has finer-grained labels, which enhances the effect of semantic guidance for feature mixup in FedSM.
When CLIP is replaced by MetaCLIP, representing a specific domain shift, the results remain close to those obtained with CLIP and are competitive with other baseline results. 
This demonstrates FedSM's robustness to domain shift between the pretrained model and training data.  

\paragraph{Results on ImageNet-LT.}  
For fine-grained analysis, we divide categories of the full ImageNet-LT dataset into three groups based on samples amounts:  
\textit{many} ($>$100 samples), \textit{medium} (20–100 samples), and \textit{few} ($<$20 samples).  
Table~\ref{table:top_one_acc_imagenet} shows the results of the \textit{overall} dataset along with divided groups.
Despite the substantial imbalance in ImageNet-LT, FedSM with CLIP and MetaCLIP achieves the overall accuracy of 30.9\% and 29.3\%, an improvement of 3.4 and 1.8 percentage points compared to the previous SOTA (27.5\%).  
Even with fewer retraining rounds, our method matches or surpasses others, particularly on tail classes (\textit{Few}) with the accuracy of 23.0\%.  
Note that FedSM achieves this performance efficiently with classifier retraining only in the final 50 communication rounds (50 epochs each), 
while CLIP2FL requires gradient matching for 300 epochs in every round throughput the training (totally 200 rounds).
These results highlight FedSM's computational efficiency and robustness under severely skewed data.


\subsection{Ablation Study}

\begin{table}[!t]
\fontsize{9}{10}\selectfont
    \centering
    \begin{tabular}{ccc}
        \toprule
        \textbf{FedSM}& \textbf{CIFAR-10} & \textbf{CIFAR-100} \\
        \midrule
        w/o SR & 71.2 & 36.0\\
        w/ deterministic SR & 70.8& 35.6 \\
        w/ probabilistic SR (Ours) & \textbf{72.2} &  \textbf{37.8} \\
        \bottomrule
    \end{tabular}
    \caption{Impact of semantic relevance (SR) on accuracy (\%).}
    \label{abl:without_rank}
\end{table}

\textbf{Effect of Semantic Relevance.}  
To assess the contribution of semantic relevance, we replace the probabilistic relevance-guided mixup mechanism with random mixup without considering relevance and deterministic mixup relying on relevance.  
As shown in Table~\ref{abl:without_rank} results under an IF of 100, removing semantic relevance leads to a performance drop of approximately 1 to 2 percentage points across both datasets, confirming its importance in guiding effective pseudo feature generation. 
While the method relying on deterministic relevance can be subject to potential domain shift biases (resulting in a more than two percentage point drop), our probabilistic method effectively addresses this issue by increasing feature diversity.

\begin{table}[!t]
\fontsize{9}{10}\selectfont
    \centering
    \begin{tabular}{cccc}
        \toprule
        \textbf{FedSM}& \textbf{IF=100} & \textbf{IF=50} & \textbf{IF=10} \\
        \midrule
        w/o fine-tuning & 37.8 & 41.2 & 50.7 \\
        w/ fine-tuning & \textbf{38.4} & \textbf{42.4} & \textbf{52.0} \\
        \bottomrule
    \end{tabular}
    \caption{Accuracy (\%) on CIFAR-100-LT with and without fine-tuning.}
    \label{abl:finetuning}
\end{table}

\textbf{Effect of Fine-tuning.}  
To investigate FedSM potential ability to mitigate the domain shift problem, we explore the effect of fine-tuning during local training. 
Instead of freezing the image encoder of CLIP, we optimize it by the loss of Margin Metric Softmax~\cite{clipood_icml23}, which adds an adaptive margin for each negative feature pair between the image and text encoders. 
Since fine-tuning the teacher model during knowledge distillation based on logits between the teacher and student models can lead to unstable training, we further replace the  Kullback–Leibler divergence between logits in Eq.~\ref{loss local learning} with the mean square error between features to optimize the feature extractor, following the practice suggested by FitNets~\cite{romero2015fitnetshintsdeepnets}. 
Table~\ref{abl:finetuning} shows the results on CIFAR-100-LT, a dataset with finer-grained categories and is more likely subject to problems caused by domain shift.
After fine-tuning, FedSM delivered notably improved accuracy across different imbalance factors, 
e.g., an extra 1.3 percentage point improvement when IF=10.
This fine-tuning helps quickly refine pretrained image feature spaces to align with out-of-domain training data, enhancing the results of data augmentation based on semantic relevance.

\begin{figure}[!t]
    \centering
    \includegraphics[width=0.5\linewidth]{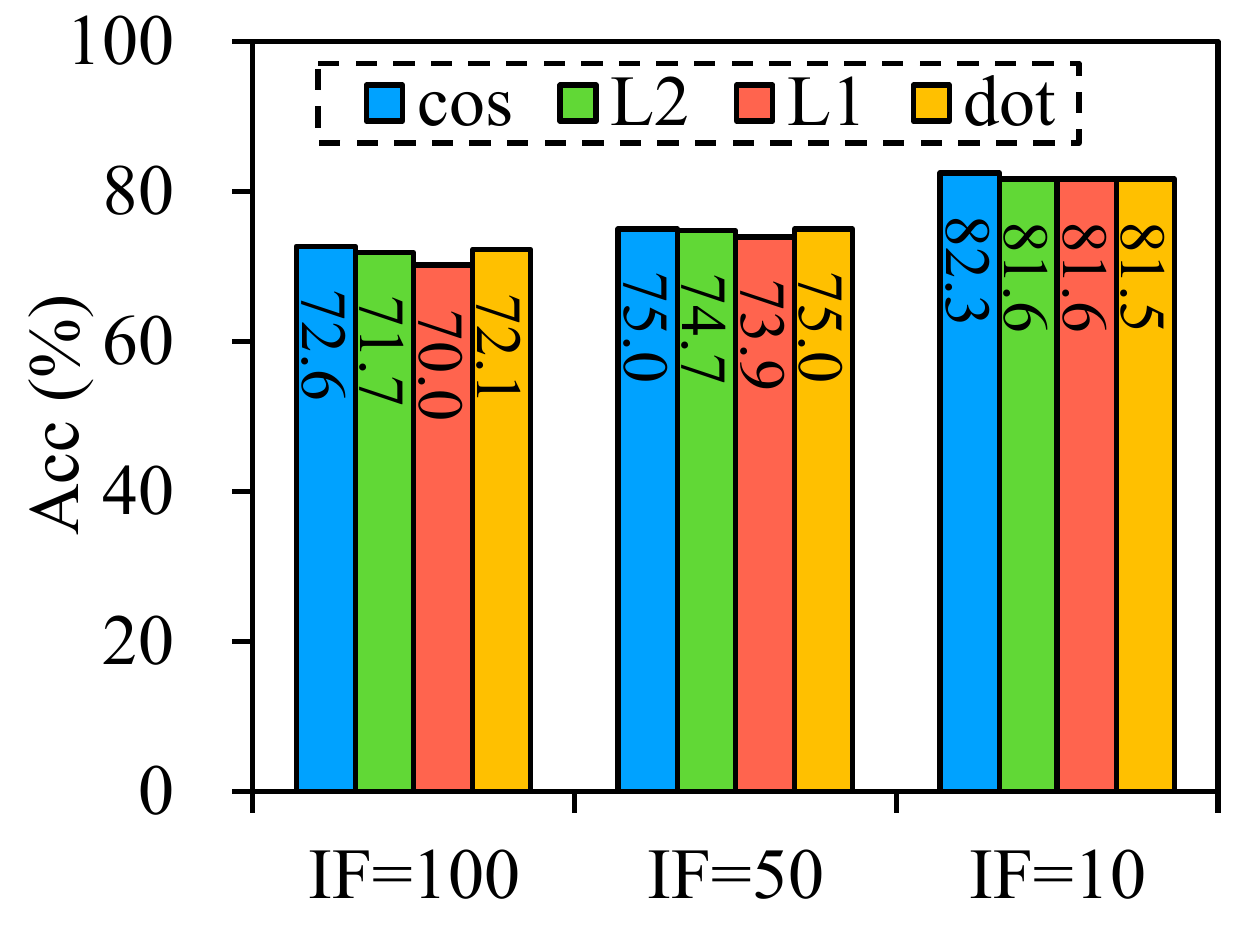}
    \caption{Results on CIFAR-10-LT under different similarity functions for relevance score.}
    \label{abl:dis_fun}
\end{figure}

\textbf{Effect of Distance Functions on Relevance Calculation.}  
We evaluate the impact of different similarity distance functions used in relevance score calculation, as defined in Eq.~\ref{eq:relevance_score}.  
Experiments on CIFAR-10-LT are conducted under various IF settings, comparing four common distance functions: cosine similarity, L2 distance, L1 distance, and dot product.
As shown in Fig.~\ref{abl:dis_fun}, cosine similarity yields the best performance across all IF levels, consistent with its widespread use in semantic similarity tasks.  
L1 and L2 distances yield lower accuracy, especially under high imbalance (IF=100), suggesting that they are less robust in capturing meaningful semantic relationships in sparse or skewed feature distributions.   
These results highlight the importance of selecting an appropriate similarity function to guide relevance-aware mixup in long-tail FL scenarios.



\begin{figure}[!t]
    \begin{subfigure}[t]{0.48\columnwidth}
        \centering
        \includegraphics[width=\columnwidth]{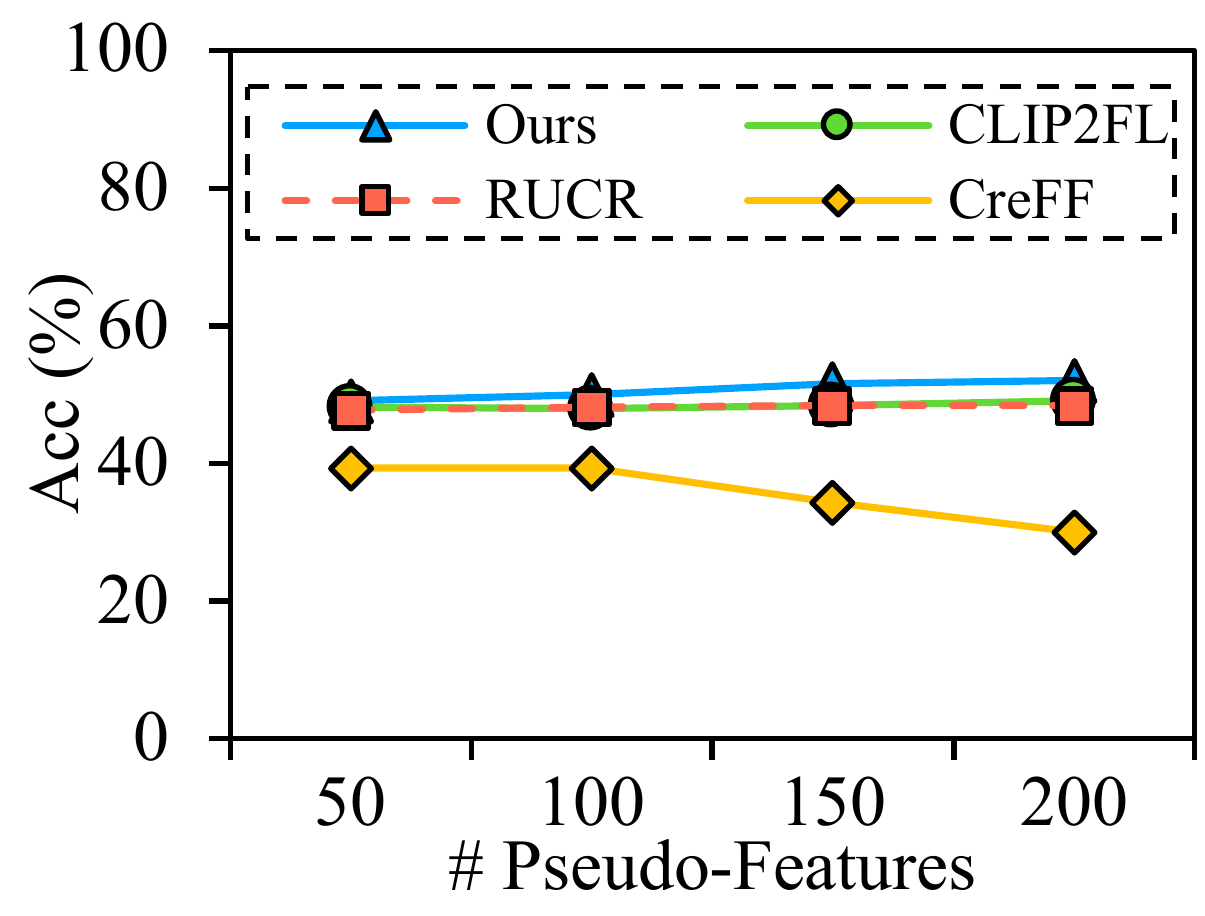}
        \caption{Varying number of pseudo-features.}
        \label{abl:num_of_features}
    \end{subfigure}
    \hspace{0.01\columnwidth}
    \begin{subfigure}[t]{0.48\columnwidth}
        \centering
        \includegraphics[width=\columnwidth]{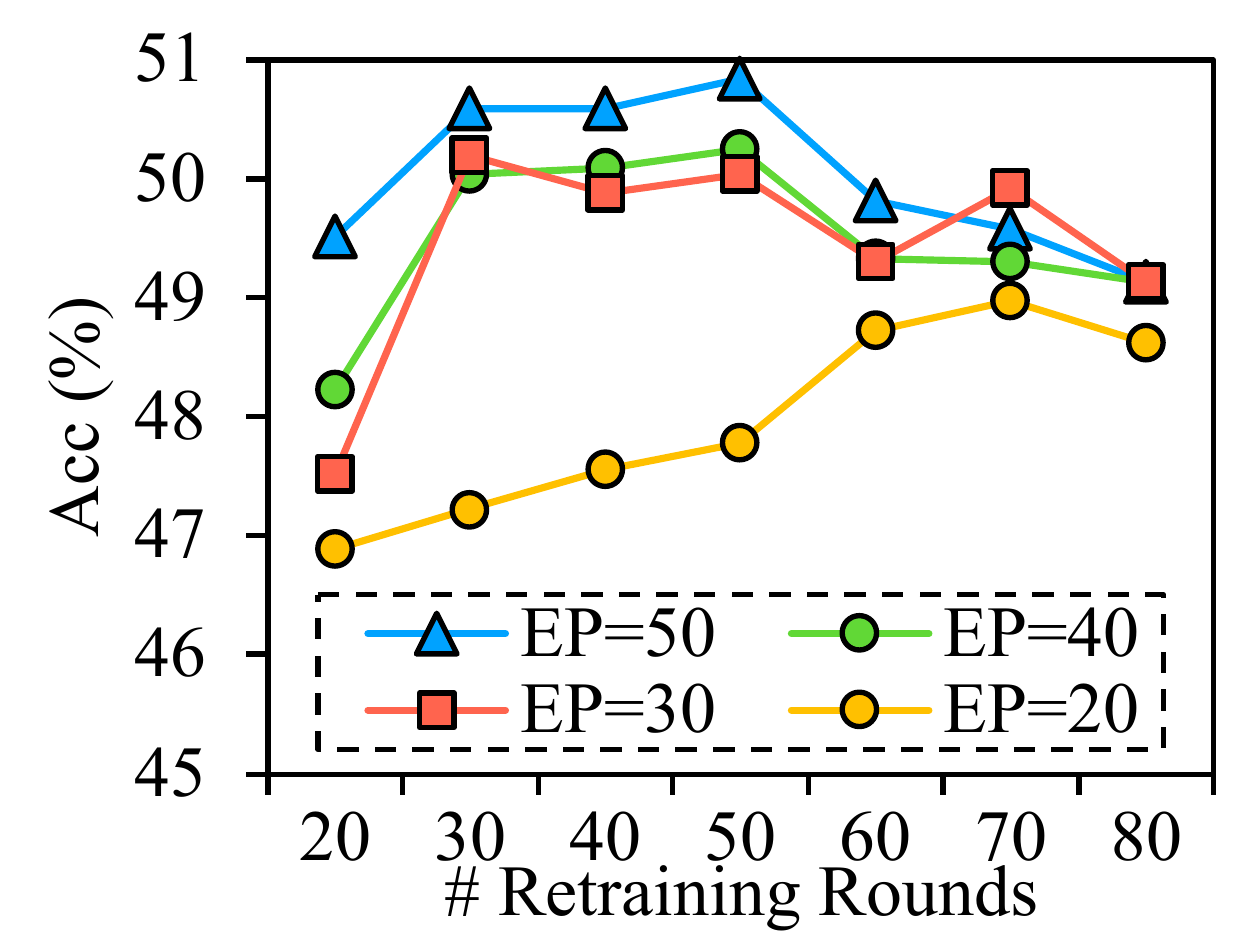}
        \caption{Varying number of retraining epochs (EP). The y axis ranges between 45\% and 51\%.}
        \label{abl:crt_ep}
    \end{subfigure}
    \caption{Results in various classifier retraining settings on CIFAR-100-LT with IF=10.}
\end{figure}

\textbf{Effect of the Number of Pseudo-features.}  
We evaluate FedSM's performance on CIFAR-100-LT with IF=10 when generating varying numbers of pseudo-features for classifier retraining, as shown in Fig.~\ref{abl:num_of_features}.  
FedSM consistently benefits from increasing the number of pseudo-features, with each additional 50 samples yielding an approximate 1 percentage point improvement in accuracy.  
This gain is not solely due to quantity, but also to more mixup operations that encourage a more uniform and balanced feature distribution, helping to reduce classifier bias and refine decision boundaries.


Interestingly, CLIP2FL and RUCR also exhibit slight performance gains with more pseudo-features, albeit at a lower speed than FedSM.  
In contrast, CReFF shows declining accuracy as the number increases.  
A possible explanation is that CReFF relies on average gradient matching to optimize pseudo-features, which may yield lower-quality samples.  
Additionally, increasing the pseudo feature count in CReFF likely exacerbates the optimization burden, hindering effective classifier retraining.

\textbf{Effect of the Number of Classifier Retraining Epochs.}  
We examine the impact of varying classifier retraining epochs on CIFAR-100 with IF=10.  
As shown in Fig.~\ref{abl:crt_ep}, FedSM achieves comparable performance using only 50 retraining rounds with 50 epochs each, limited to the final phase of training.  
In contrast, prior methods~\cite{clip2fl, CReFF} perform retraining in every communication round with 300 epochs, leading to significantly higher computational costs.  
This highlights the efficiency of our approach in reducing training overhead without sacrificing accuracy.

\begin{figure}[!t]
    \centering
    \begin{subfigure}[t]{0.48\columnwidth}
        \centering
        \includegraphics[width=\textwidth]{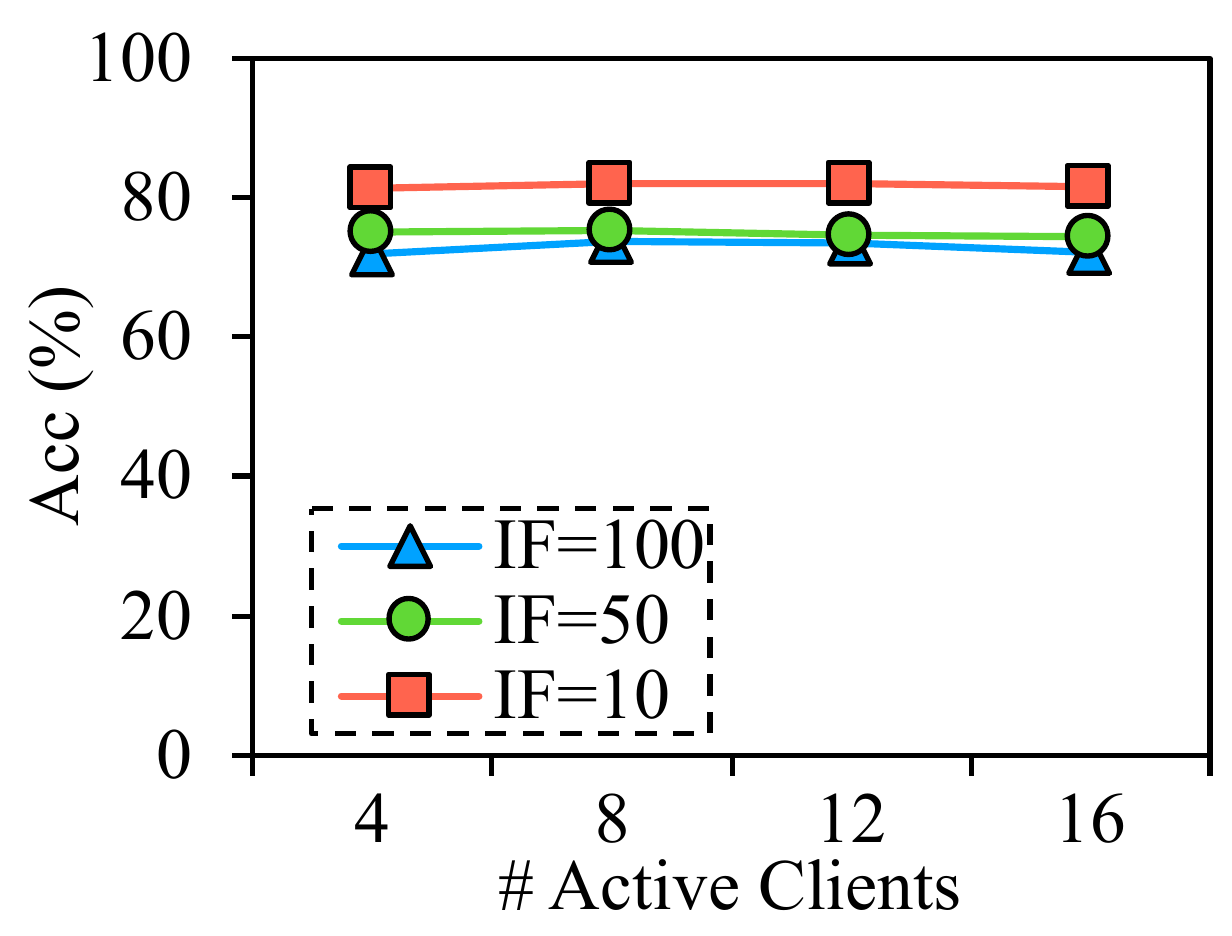}
        \centerline{(a) CIFAR-10-LT}
    \end{subfigure}
    \begin{subfigure}[t]{0.48\columnwidth}
        \centering
        \includegraphics[width=\textwidth]{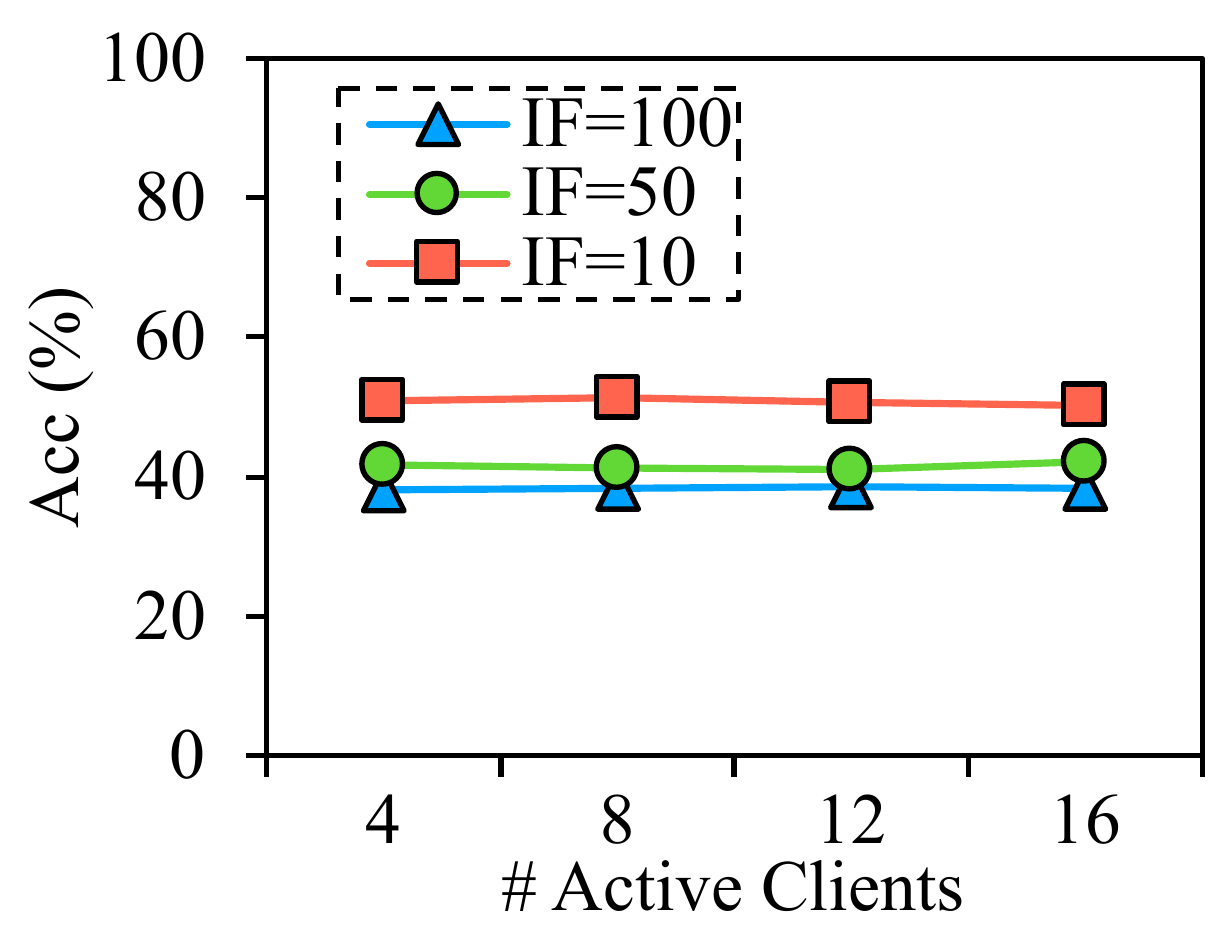}
        \centerline{(b) CIFAR-100-LT}
    \end{subfigure}
    \caption{Impact of varying the number of active clients.}
    \label{abl:active_clients_number}
\end{figure}

\textbf{Effect of the Number of Active Clients.}  
We evaluate FedSM's performance with different numbers of active clients, a key factor in FL. 
As shown in Fig.~\ref{abl:active_clients_number}, FedSM demonstrates strong robustness to the number of active clients.  
Performance on CIFAR-10 exhibits slightly more fluctuation than on CIFAR-100, possibly due to less distinct label semantics in CIFAR-10.  
Across all settings, lower imbalance (i.e., smaller IF values) consistently yields higher accuracy, which aligns with trends observed in the main results.


\begin{table}[!t]
\fontsize{9}{10}\selectfont
\centering
    \begin{tabular}{ccccccc}
        \toprule
           & \multicolumn{3}{c}{\textbf{CIFAR-10-LT}} & \multicolumn{3}{c}{\textbf{CIFAR-100-LT}} \cr\cline{2-4} \cline{5-7}
           & IF=100 & IF=50 & IF=10 & IF=100 & IF=50 & IF=10 \\ 
        \midrule
            $\lambda$=0.20  & 71.4 & 73.6 & 81.2 & 36.0 & 39.9 & 50.0 \\
            $\lambda$=0.35  & 71.6 & 74.2 & 81.5 & 36.0 & 39.9 & 50.2 \\
            $\lambda$=0.50  & 71.7 & 74.6 & 81.5 & 36.0 & 39.9 & 50.0 \\
            $\lambda$=0.65  & 72.1 & 75.0 & 81.8 & 37.4 & 41.1 & 49.1 \\
            $\lambda$=0.80  & \textbf{72.2} & \textbf{75.1} & \textbf{81.9} & \textbf{38.0} & \textbf{41.5} & \textbf{49.4} \\
        \bottomrule
    \end{tabular}
\caption{Accuracy (\%) under varying hyperparameter $\lambda$.}
\label{table:lambda}
\end{table}

\textbf{Hyperparameter for Pseudo Feature Mixup.}  
We study the effect of the mixup coefficient $\lambda$ in Eq.~\ref{eq:mix}, which controls the interpolation between the global prototype and local feature.  
As shown in Table~\ref{table:lambda}, FedSM performs robustly across a range of $\lambda$ values from 0.20 to 0.80 with a gap of 0.15.  
Performance slightly improves as $\lambda$ increases, suggesting that stronger alignment with the global prototype enhances the semantic clarity of generated pseudo-features, thereby improving classification accuracy.

\section{Conclusion}
This paper introduces FedSM, a semantics-guided mixup framework to address classification bias in FL with long-tail data. 
FedSM leverages a pretrained image-text-aligned model to guide feature-level mixup between local features and global prototypes, generating balanced pseudo-features for few-round classifier retraining. 
It can effectively mitigate the domain shift problems via several mechanisms.
All procedures are executed locally, preserving data privacy and reducing computational overhead. 
Extensive experiments demonstrate that FedSM outperforms prior SOTA methods in accuracy, robustness to domain shift, and computational efficiency, especially in heavily skewed cases.

\bibliography{main}

\begin{thebibliography}{38}
\providecommand{\natexlab}[1]{#1}

\bibitem[{Carratino et~al.(2022)Carratino, Ciss{\'e}, Jenatton, and
  Vert}]{mixupRegularization2022JMLR}
Carratino, L.; Ciss{\'e}, M.; Jenatton, R.; and Vert, J.-P. 2022.
\newblock On mixup regularization.
\newblock \emph{Journal of machine learning research}, 23(325): 1--31.

\bibitem[{Chen and Chao(2021)}]{chen2021fedbe}
Chen, H.-Y.; and Chao, W.-L. 2021.
\newblock FedBE: Making bayesian model ensemble applicable to federated
  learning.
\newblock In \emph{ICLR}.

\bibitem[{Chen et~al.(2023)Chen, Zhou, Wu, Yang, Li, Hu, and Wang}]{area}
Chen, X.; Zhou, Y.; Wu, D.; Yang, C.; Li, B.; Hu, Q.; and Wang, W. 2023.
\newblock Area: adaptive reweighting via effective area for long-tailed
  classification.
\newblock In \emph{Proceedings of the IEEE/CVF International Conference on
  Computer Vision}, 19277--19287.

\bibitem[{Cui et~al.(2019)Cui, Jia, Lin, Song, and Belongie}]{effnumber}
Cui, Y.; Jia, M.; Lin, T.-Y.; Song, Y.; and Belongie, S. 2019.
\newblock Class-Balanced Loss Based on Effective Number of Samples.
\newblock In \emph{Proceedings of the IEEE/CVF Conference on Computer Vision
  and Pattern Recognition (CVPR)}.

\bibitem[{Duan et~al.(2020)Duan, Liu, Chen, Liu, Tan, and Liang}]{duan2020self}
Duan, M.; Liu, D.; Chen, X.; Liu, R.; Tan, Y.; and Liang, L. 2020.
\newblock Self-balancing federated learning with global imbalanced data in
  mobile systems.
\newblock \emph{IEEE Transactions on Parallel and Distributed Systems}, 32(1):
  59--71.

\bibitem[{He et~al.(2016)He, Zhang, Ren, and Sun}]{resnet}
He, K.; Zhang, X.; Ren, S.; and Sun, J. 2016.
\newblock Deep residual learning for image recognition.
\newblock In \emph{Proceedings of the IEEE conference on computer vision and
  pattern recognition}, 770--778.

\bibitem[{Hsu, Qi, and Brown(2019)}]{fedavgm}
Hsu, T.-M.~H.; Qi, H.; and Brown, M. 2019.
\newblock Measuring the effects of non-identical data distribution for
  federated visual classification.
\newblock \emph{arXiv preprint arXiv:1909.06335}.

\bibitem[{Huang et~al.(2021)Huang, Shang, Liu, and Liu}]{huang2021behavior}
Huang, H.; Shang, F.; Liu, Y.; and Liu, H. 2021.
\newblock Behavior mimics distribution: Combining individual and group
  behaviors for federated learning.
\newblock In \emph{IJCAI}, 2556--2552.

\bibitem[{Huang et~al.(2024)Huang, Liu, Ye, Chen, and Du}]{rucr}
Huang, W.; Liu, Y.; Ye, M.; Chen, J.; and Du, B. 2024.
\newblock Federated Learning with Long-Tailed Data via Representation
  Unification and Classifier Rectification.
\newblock \emph{IEEE Transactions on Information Forensics and Security}.

\bibitem[{Joyce(2011)}]{joyce2011KL}
Joyce, J.~M. 2011.
\newblock Kullback-leibler divergence.
\newblock In \emph{International encyclopedia of statistical science},
  720--722. Springer.

\bibitem[{Kang et~al.(2019)Kang, Xie, Rohrbach, Yan, Gordo, Feng, and
  Kalantidis}]{kang2019decoupling}
Kang, B.; Xie, S.; Rohrbach, M.; Yan, Z.; Gordo, A.; Feng, J.; and Kalantidis,
  Y. 2019.
\newblock Decoupling representation and classifier for long-tailed recognition.
\newblock \emph{arXiv preprint arXiv:1910.09217}.

\bibitem[{Krizhevsky, Hinton et~al.(2009)}]{cifar}
Krizhevsky, A.; Hinton, G.; et~al. 2009.
\newblock Learning multiple layers of features from tiny images.

\bibitem[{Li, He, and Song(2021)}]{moon}
Li, Q.; He, B.; and Song, D. 2021.
\newblock Model-Contrastive Federated Learning.
\newblock In \emph{Proceedings of the IEEE/CVF Conference on Computer Vision
  and Pattern Recognition (CVPR)}, 10713--10722.

\bibitem[{Li et~al.(2020{\natexlab{a}})Li, Sahu, Zaheer, Sanjabi, Talwalkar,
  and Smith}]{MLSYS2020_38af8613}
Li, T.; Sahu, A.~K.; Zaheer, M.; Sanjabi, M.; Talwalkar, A.; and Smith, V.
  2020{\natexlab{a}}.
\newblock Federated optimization in heterogeneous networks.
\newblock In \emph{MLSys}, 429--450.

\bibitem[{Li et~al.(2020{\natexlab{b}})Li, Sahu, Zaheer, Sanjabi, Talwalkar,
  and Smith}]{fedprox}
Li, T.; Sahu, A.~K.; Zaheer, M.; Sanjabi, M.; Talwalkar, A.; and Smith, V.
  2020{\natexlab{b}}.
\newblock Federated optimization in heterogeneous networks.
\newblock \emph{Proceedings of Machine learning and systems}, 2: 429--450.

\bibitem[{Liu et~al.(2019)Liu, Miao, Zhan, Wang, Gong, and Yu}]{iamgenet-lt}
Liu, Z.; Miao, Z.; Zhan, X.; Wang, J.; Gong, B.; and Yu, S.~X. 2019.
\newblock Large-scale long-tailed recognition in an open world.
\newblock In \emph{Proceedings of the IEEE/CVF conference on computer vision
  and pattern recognition}, 2537--2546.

\bibitem[{Luo et~al.(2021)Luo, Chen, Hu, Zhang, Liang, and Feng}]{ccvr}
Luo, M.; Chen, F.; Hu, D.; Zhang, Y.; Liang, J.; and Feng, J. 2021.
\newblock No fear of heterogeneity: Classifier calibration for federated
  learning with non-iid data.
\newblock \emph{Advances in Neural Information Processing Systems}, 34:
  5972--5984.

\bibitem[{McMahan et~al.(2017{\natexlab{a}})McMahan, Moore, Ramage, Hampson,
  and Arcas}]{FL}
McMahan, B.; Moore, E.; Ramage, D.; Hampson, S.; and Arcas, B. A.~y.
  2017{\natexlab{a}}.
\newblock {Communication-Efficient Learning of Deep Networks from Decentralized
  Data}.
\newblock In Singh, A.; and Zhu, J., eds., \emph{Proceedings of the 20th
  International Conference on Artificial Intelligence and Statistics},
  volume~54 of \emph{Proceedings of Machine Learning Research}, 1273--1282.
  PMLR.

\bibitem[{McMahan et~al.(2017{\natexlab{b}})McMahan, Moore, Ramage, Hampson,
  and y~Arcas}]{fedavg}
McMahan, B.; Moore, E.; Ramage, D.; Hampson, S.; and y~Arcas, B.~A.
  2017{\natexlab{b}}.
\newblock Communication-efficient learning of deep networks from decentralized
  data.
\newblock In \emph{Artificial intelligence and statistics}, 1273--1282. PMLR.

\bibitem[{Radford et~al.(2021)Radford, Kim, Hallacy, Ramesh, Goh, Agarwal,
  Sastry, Askell, Mishkin, Clark et~al.}]{clip}
Radford, A.; Kim, J.~W.; Hallacy, C.; Ramesh, A.; Goh, G.; Agarwal, S.; Sastry,
  G.; Askell, A.; Mishkin, P.; Clark, J.; et~al. 2021.
\newblock Learning transferable visual models from natural language
  supervision.
\newblock In \emph{International conference on machine learning}, 8748--8763.
  PMLR.

\bibitem[{Romero et~al.(2015)Romero, Ballas, Kahou, Chassang, Gatta, and
  Bengio}]{romero2015fitnetshintsdeepnets}
Romero, A.; Ballas, N.; Kahou, S.~E.; Chassang, A.; Gatta, C.; and Bengio, Y.
  2015.
\newblock FitNets: Hints for Thin Deep Nets.
\newblock arXiv:1412.6550.

\bibitem[{Russakovsky et~al.(2015)Russakovsky, Deng, Su, Krause, Satheesh, Ma,
  Huang, Karpathy, Khosla, Bernstein, Berg, and Fei-Fei}]{ImageNet}
Russakovsky, O.; Deng, J.; Su, H.; Krause, J.; Satheesh, S.; Ma, S.; Huang, Z.;
  Karpathy, A.; Khosla, A.; Bernstein, M.; Berg, A.~C.; and Fei-Fei, L. 2015.
\newblock {ImageNet Large Scale Visual Recognition Challenge}.
\newblock \emph{International Journal of Computer Vision (IJCV)}, 115(3):
  211--252.

\bibitem[{Sarkar, Narang, and Rai(2020)}]{fedfocalloss}
Sarkar, D.; Narang, A.; and Rai, S. 2020.
\newblock Fed-focal loss for imbalanced data classification in federated
  learning.
\newblock \emph{arXiv preprint arXiv:2011.06283}.

\bibitem[{Shang et~al.(2022)Shang, Lu, Huang, and Wang}]{CReFF}
Shang, X.; Lu, Y.; Huang, G.; and Wang, H. 2022.
\newblock Federated learning on heterogeneous and long-tailed data via
  classifier re-training with federated features.
\newblock \emph{arXiv preprint arXiv:2204.13399}.

\bibitem[{Shi et~al.(2024)Shi, Zheng, Yin, Lu, Xie, and Qu}]{clip2fl}
Shi, J.; Zheng, S.; Yin, X.; Lu, Y.; Xie, Y.; and Qu, Y. 2024.
\newblock CLIP-Guided Federated Learning on Heterogeneity and Long-Tailed Data.
\newblock In \emph{Proceedings of the AAAI Conference on Artificial
  Intelligence}, volume~38, 14955--14963.

\bibitem[{Shokri and Shmatikov(2015)}]{shokri2015privacy}
Shokri, R.; and Shmatikov, V. 2015.
\newblock Privacy-preserving deep learning.
\newblock In \emph{Proceedings of the 22nd ACM SIGSAC conference on computer
  and communications security}, 1310--1321.

\bibitem[{Shu et~al.(2023)Shu, Guo, Wu, Wang, Wang, and Long}]{clipood_icml23}
Shu, Y.; Guo, X.; Wu, J.; Wang, X.; Wang, J.; and Long, M. 2023.
\newblock {CLIP}ood: Generalizing {CLIP} to Out-of-Distributions.
\newblock In \emph{Proceedings of the 40th International Conference on Machine
  Learning}, volume 202, 31716--31731. PMLR.

\bibitem[{Teney, Wang, and Abbasnejad(2024)}]{selectiveMixup2024icml}
Teney, D.; Wang, J.; and Abbasnejad, E. 2024.
\newblock Selective Mixup Helps with Distribution Shifts, But Not (Only)
  because of Mixup.
\newblock In \emph{International Conference on Machine Learning}, 47948--47964.

\bibitem[{Wang et~al.(2020)Wang, Liu, Liang, Joshi, and Poor}]{fednova}
Wang, J.; Liu, Q.; Liang, H.; Joshi, G.; and Poor, H.~V. 2020.
\newblock Tackling the objective inconsistency problem in heterogeneous
  federated optimization.
\newblock \emph{Advances in neural information processing systems}, 33:
  7611--7623.

\bibitem[{Wang et~al.(2021{\natexlab{a}})Wang, Xu, Wang, and
  Zhu}]{fedratioloss}
Wang, L.; Xu, S.; Wang, X.; and Zhu, Q. 2021{\natexlab{a}}.
\newblock Addressing class imbalance in federated learning.
\newblock In \emph{Proceedings of the AAAI Conference on Artificial
  Intelligence}, volume~35, 10165--10173.

\bibitem[{Wang et~al.(2023)Wang, Fan, Peng, Li, Yang, Feng, Yang, Liu, and
  Wang}]{wangzheng2023flgo}
Wang, Z.; Fan, X.; Peng, Z.; Li, X.; Yang, Z.; Feng, M.; Yang, Z.; Liu, X.; and
  Wang, C. 2023.
\newblock FLGo: A Fully Customizable Federated Learning Platform.
\newblock arXiv:2306.12079.

\bibitem[{Wang et~al.(2021{\natexlab{b}})Wang, Fan, Qi, Wen, Wang, and
  Yu}]{wangzheng2021federated}
Wang, Z.; Fan, X.; Qi, J.; Wen, C.; Wang, C.; and Yu, R. 2021{\natexlab{b}}.
\newblock Federated Learning with Fair Averaging.
\newblock arXiv:2104.14937.

\bibitem[{Xu et~al.()Xu, Xie, Tan, Huang, Howes, Sharma, Li, Ghosh,
  Zettlemoyer, and Feichtenhofer}]{metaclip_iclr24}
Xu, H.; Xie, S.; Tan, X.; Huang, P.-Y.; Howes, R.; Sharma, V.; Li, S.-W.;
  Ghosh, G.; Zettlemoyer, L.; and Feichtenhofer, C. ????
\newblock Demystifying CLIP Data.
\newblock In \emph{The Twelfth International Conference on Learning
  Representations}.

\bibitem[{Yang et~al.(2021)Yang, Wang, Zhu, Wang, and Qian}]{yang2021federated}
Yang, M.; Wang, X.; Zhu, H.; Wang, H.; and Qian, H. 2021.
\newblock Federated learning with class imbalance reduction.
\newblock In \emph{2021 29th European Signal Processing Conference (EUSIPCO)},
  2174--2178. IEEE.

\bibitem[{Yun et~al.(2019)Yun, Han, Oh, Chun, Choe, and Yoo}]{yun2019cutmix}
Yun, S.; Han, D.; Oh, S.~J.; Chun, S.; Choe, J.; and Yoo, Y. 2019.
\newblock Cutmix: Regularization strategy to train strong classifiers with
  localizable features.
\newblock In \emph{Proceedings of the IEEE/CVF international conference on
  computer vision}, 6023--6032.

\bibitem[{Zhang et~al.(2018)Zhang, Cisse, Dauphin, and
  Lopez-Paz}]{iclr2018mixup}
Zhang, H.; Cisse, M.; Dauphin, Y.~N.; and Lopez-Paz, D. 2018.
\newblock mixup: Beyond Empirical Risk Minimization.
\newblock In \emph{International Conference on Learning Representations}.

\bibitem[{Zhang et~al.(2025)Zhang, Liang, Yuan, Yang, Li, and
  Hu}]{fedpall_iccv25}
Zhang, Y.; Liang, F.; Yuan, G.; Yang, M.; Li, C.; and Hu, X. 2025.
\newblock FedPall: Prototype-based Adversarial and Collaborative Learning for
  Federated Learning with Feature Drift.
\newblock In \emph{International Conference on Computer Vision}, 1 -- 10.

\bibitem[{Zhou et~al.(2020)Zhou, Cui, Wei, and Chen}]{zhou2020bbn}
Zhou, B.; Cui, Q.; Wei, X.-S.; and Chen, Z.-M. 2020.
\newblock Bbn: Bilateral-branch network with cumulative learning for
  long-tailed visual recognition.
\newblock In \emph{Proceedings of the IEEE/CVF conference on computer vision
  and pattern recognition}, 9719--9728.

\end{thebibliography}
\def\isChecklistMainFile{true}

\end{document}